\documentclass{article}

\usepackage[accepted]{icml2021}
\usepackage{microtype, graphicx, subfigure, booktabs, bbold, enumitem, subfigure, times, soul, url, graphicx, amsmath, amsthm}
\usepackage[hidelinks]{hyperref}
\usepackage[utf8]{inputenc}
\usepackage[small]{caption}

\icmltitlerunning{GRP-FED: Addressing Client Imbalance in Federated Learning via Global-Regularized Personalization}

\begin{document}

\twocolumn[
\icmltitle{GRP-FED: Addressing Client Imbalance in Federated Learning via Global-Regularized Personalization}
\icmlsetsymbol{equal}{*}
\begin{icmlauthorlist}
    \icmlauthor{Yen-Hsiu Chou}{pku_cs,pku_me}
    \icmlauthor{Shenda Hong}{pku_hd,pku_hs}
    \icmlauthor{Chenxi Sun}{pku_cs,pku_me}
    \icmlauthor{Derun Cai}{pku_cs,pku_me}
    \icmlauthor{Moxian Song}{pku_cs,pku_me}
    \icmlauthor{Hongyan Li}{pku_cs,pku_me}
\end{icmlauthorlist}
\icmlaffiliation{pku_cs}{School of Electronics Engineering and Computer Science, Peking University, Beijing, People’s Republic of China}
\icmlaffiliation{pku_hd}{National Institute of Health Data Science, Peking University, Beijing,  People’s Republic of China}
\icmlaffiliation{pku_me}{Key Laboratory of Machine Perception (Ministry of Education), Peking University, Beijing,  People’s Republic of China}
\icmlaffiliation{pku_hs}{Institute of Medical Technology, Health Science Center of Peking University, Beijing,  People’s Republic of China}
\icmlcorrespondingauthor{Yen-Hsiu Chou}{emily051485@gmail.com}
\icmlkeywords{federated learning}

\vskip 0.3in
]
\printAffiliationsAndNotice{} 

\begin{abstract}
Since data is presented long-tailed in reality, it is challenging for Federated Learning (FL) to train across decentralized clients as practical applications. We present Global-Regularized Personalization (GRP-FED) to tackle the data imbalanced issue by considering a single global model and multiple local models for each client. With adaptive aggregation, the global model treats multiple clients fairly and mitigates the global long-tailed issue. Each local model is learned from the local data and aligns with its distribution for customization. To prevent the local model from just overfitting, GRP-FED applies an adversarial discriminator to regularize between the learned global-local features. Extensive results show that our GRP-FED improves under both global and local scenarios on real-world MIT-BIH and synthesis CIFAR-10 datasets, achieving comparable performance and addressing client imbalance. 
\end{abstract}

\section{Introduction}
Federated Learning (FL) is a distributed learning algorithm that trains models across multiple decentralized clients and keeps data private simultaneously \cite{DBLP:conf/aistats/McMahanMRHA17,konevcny2016federated}. One of the issues in FL is the distinct distributions where decentralized data is diverse due to different properties over each client \cite{kairouz2019advances,liang2020think,deng2020adaptive,li2019fair}. 

\begin{figure}[t]
    \centering
    \includegraphics[width=\linewidth]{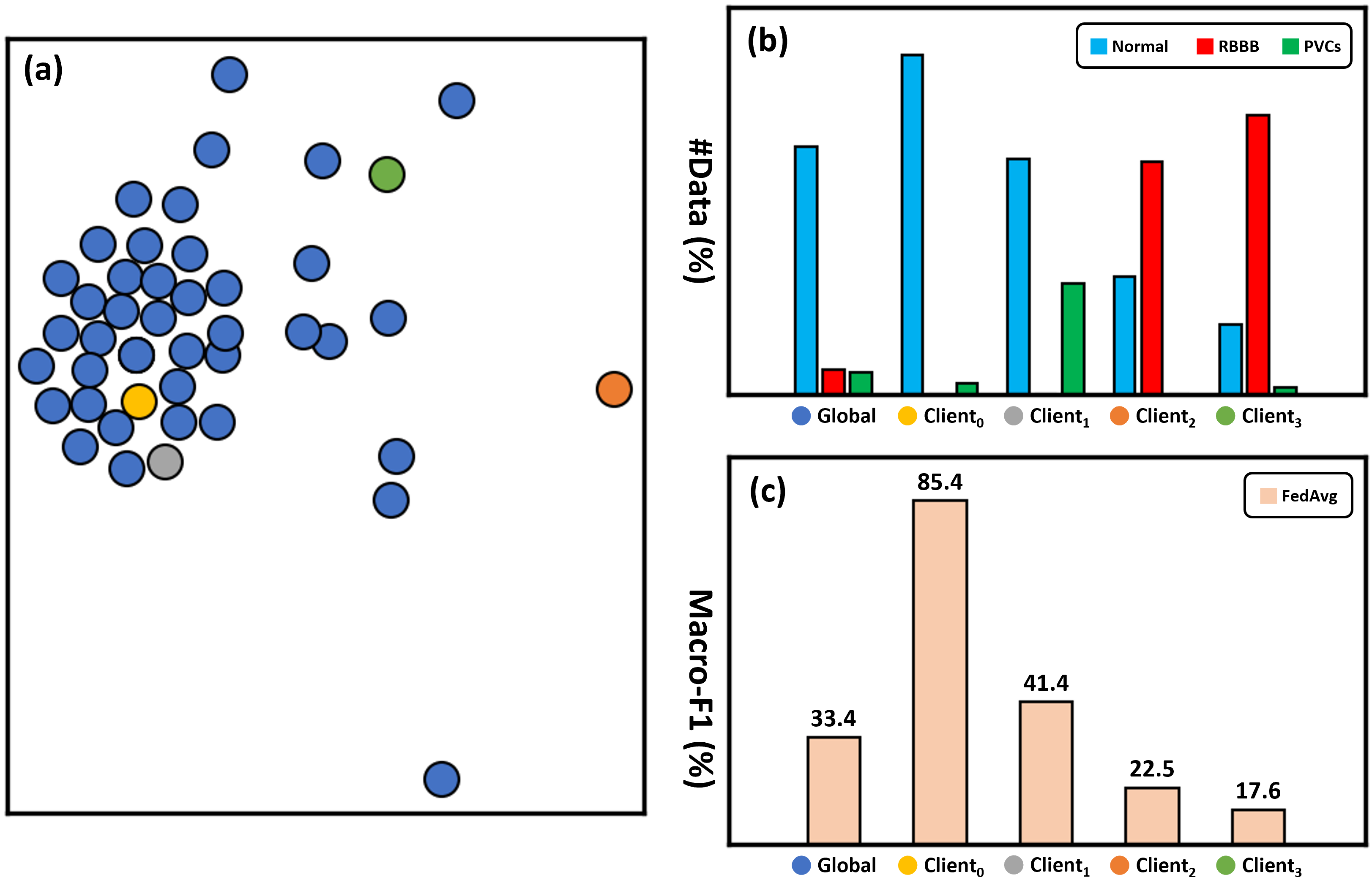}
    \vspace{-3ex}
    \caption{The client imbalanced issue in the real-world MIT-BIH dataset. (a) visualizes the client distribution as each client point via t-SNE; (b) plots the class distribution of the global and selected clients; (c) presents the performance (Macro-F1) of FedAvg on the entire test (global) or specific clients (local).}
    \vspace{-3ex}
    \label{fig:Intro}
\end{figure}

In the real world, data is inevitably long-tailed \cite{DBLP:conf/nips/YangX20}. Fig.~\ref{fig:Intro} illustrates the data distribution of MIT-BIH \cite{goldberger2000physiobank}, a real-world Electrocardiography (ECG) dataset for medical diagnosis. Each patient, which may have different arrhythmia issues over several ECGs, is viewed as a client under the FL setting. Fig.~\ref{fig:Intro}(a) visualizes the client distribution as each client point using t-SNE \cite{maaten2008visualizing}. It shows that the global distribution of clients is non-uniformly, where some clients are scattered and far away. Fig.~\ref{fig:Intro}(b) plots the class distribution of data from clients, which shares distinct distributions and provides different class data.

FedAvg \cite{DBLP:conf/aistats/McMahanMRHA17} is a classic FL algorithm where a single model tries to fit among clients by averaging the parameters from local training. Fig.~\ref{fig:Intro}(c) presents the Macro-F1 score of FedAvg for global and client-based (local) testing. Since conducting all clients equally, FedAvg ignores the various data distributions between clients, making the poor performance on the global test. Furthermore, FedAvg is easily dominated by major clients but gives up remaining clients, where the F1 score drops drastically on them. Fig.~\ref{fig:Intro} shows this imbalanced issue that makes applying FL to practical applications challenging. 

Even if encouraging worse clients to focus on global fairness \cite{DBLP:conf/icml/MohriSS19,li2019fair}, the performance gap between global and local tests is still significant \cite{DBLP:journals/corr/abs-1909-12488}, which indicates that personalization is crucial in FL. The local training \cite{fallah2020personalized,DBLP:conf/nips/KhodakBT19,liang2020think,DBLP:conf/nips/DinhTN20} adopts personalization by training part of local models only on client data as customization. However, the local models, which directly minimize the local error, suffer from overfitting and lose the discrimination of those minor local classes.

\begin{figure*}[t]
    \centering
    \includegraphics[width=.78\linewidth]{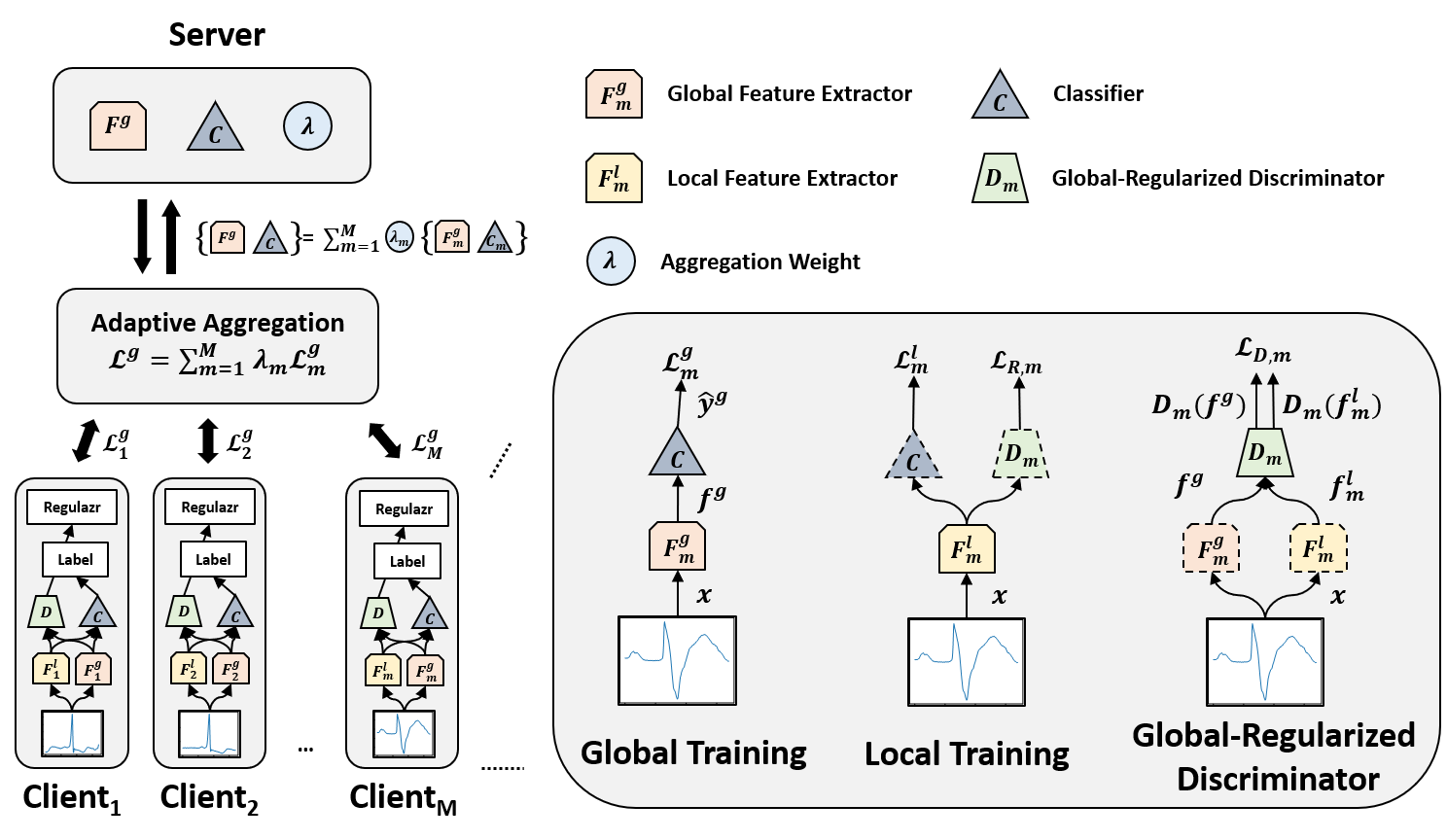}
    \vspace{-1ex}
    \caption{An overview of Global-Regularized Personalization (GRP-FED) for the federated learning (FL). For better global fiarness, we adopt adaptive aggregation to investigate different aspects and proportions of clients. We make each local client optimize only on their client to support personalization. In addition, the proposed global-regularized discriminator helps to prevent from overfitting.}
    \vspace{-2ex}
    \label{fig:Framework}
\end{figure*}

In this paper, we introduce Global-Regularized Personalization (GRP-FED) to address client imbalance in FL. As shown in Fig.~\ref{fig:Framework}, GRP-FED contains a single global model and local models for each client to consider global fairness and local personalization. Since each client provides different amounts and aspects of class data, our GRP-FED presents Adaptive Aggregation to adaptively adjust the weight of each client and aggregate as a fairer global model. To do personalization, local models are only trained on the specific data for each client. In case of being customizing but overfitting, we present the Global-Regularized Discriminator (D) to distinguish that an extracted feature is from the global or the local model. By jointly optimizing to fool D, local models learn the specific distribution for each client and the general global feature to avoid overfitting.

We conduct the evaluation on real-world MIT-BIH \cite{goldberger2000physiobank} and synthesis CIFAR-10 \cite{cifar10} datasets under FL setting. The experimental results show that our GRP-FED can improve both global and local tests. Furthermore, the proposed global-regularized discriminator addresses local overfitting effectively. In summary, our contributions are three-fold:
\begin{itemize}[topsep=0pt, noitemsep, leftmargin=*]
    \item We present GRP-FED to simultaneously consider global fairness and local personalization for Federated Learning;
    \item The proposed adaptively-aggregated global model and customized local models gain improvement under both global and local scenarios;
    \item Extensive ablation studies on both real-world MIT-BIH and synthesis CIFAR-10 show that GRP-FED achieves better performance and deals with client imbalance.
\end{itemize}

\section{Related Work}
\vspace{0.5ex} \textbf{Federated Learning (FL)~}
Federated Learning (FL) \cite{DBLP:conf/aistats/McMahanMRHA17,konevcny2016federated}, where models are trained across multiple decentralized clients, aims to preserve user privacy \cite{DBLP:conf/icml/LiKQR20,DBLP:conf/nips/0001DKD19} and lower communication costs \cite{DBLP:conf/icml/LiKQR20,DBLP:conf/nips/0001DKD19}. Similar to imbalanced data distribution \cite{DBLP:journals/corr/abs-2002-05516,DBLP:conf/nips/KhodakBT19,wang2021addressing,duan2020self}, we investigate the global model used for all and new data and local models that are customized and support personalization for local clients.

\vspace{0.5ex} \textbf{Global Model for FL~}
In FL, the global model is trained from all clients and fit the overall global distribution. FedAvg \cite{DBLP:conf/aistats/McMahanMRHA17} is the first to apply local SGD and build a single global model from a subset of clients. Moreover, they improve global fairness by adapting the global model better to each client \cite{fallah2020personalized,DBLP:conf/nips/KhodakBT19} or treating clients with different importance weights \cite{DBLP:conf/icml/MohriSS19}. Inspired by q-FFL \cite{li2019fair}, which utilizes a constant power to tune the amount of fairness, our GRP-FED adaptively adjusts the power of loss to satisfy dynamic fairness during the global training.

\vspace{0.5ex} \textbf{Local Model for FL~}
The performance gap between global and local tests indicates that personalization is crucial in FL \cite{DBLP:journals/corr/abs-1909-12488}. Local fine-tuning \cite{smith2017federated,9207508,DBLP:conf/nips/KhodakBT19,fallah2020personalized,liang2020think} supports personalization by training each local model only on client data. While, pFedMe \cite{DBLP:conf/nips/DinhTN20} argues that directly minimizing local error is prone to overfit and adopts Moreau envelopes to help decouple personalization. Different from that, our GRPFED introduces the Global-Regularized Discriminator to regularize the local feature distribution and mitigate the local overfitting issue.

\section{Approach}
\subsection{Overview}
\vspace{0.5ex} \textbf{Task Definition~}
Federated Learning (FL) is to learn from independent $M$ clients where client $m$ contains a local dataset $\mathcal{D}_m=\{(x_0, y_0)_m, ..., (x_{N_m}, y_{N_m})_m\}$. $(x_i, y_i)_m$ represents the pair of $i$th data and its label where $N_m$ is the number of data in $\mathcal{D}_m$. We consider FL as classification task where $y$ is the class label of $x$. Intuitively, each client captures a different view $p(X, Y)_m$ of the global data distribution $p(X, Y)$. However, since each client provides $\mathcal{D}$ with distinct distributions in real world, FL easily suffers from the client imbalance under practical applications. 

\vspace{0.5ex} \textbf{GRP-FED~}
To address the client imbalanced issue, we present Global-Regularized Personalization (GRP-FED) into FL. An overview of GRP-FED\footnote{$\theta^g_{F, m}$: global feature extractor, $\theta^l_{F, m}$: local feature extractor, $\theta_C$: classifier, and $\theta_{D, m}$: discriminator in client $m$.} is illustrated in Fig.~\ref{fig:Framework}. For a data point $x$, the feature extractor $F$ extracts the lower-dimensional representation of $x$, and the classifier $C$ performs the output prediction $\hat{y}$. GRP-FED consists of a fair global model that applies adaptive aggregation to consider different aspects from clients, and local models to customize each client. The proposed adaptive aggregation adjusts the aggregated proportion to ensure the fairness over distinct clients. The local models do personalization, where a global-regularized discriminator prevents it from overfitting when optimizing the client data. 

\subsection{Global Fairness}
\vspace{0.5ex} \textbf{Global Training~}
Global fairness aims at building a global model that can fairly cope with the global distribution over distinct clients. The global model includes the global feature extractor $F$ parameterized by $\theta^g_F$ and the classifier $C$ parameterized by $\theta_C$. The global training is to train the global model in each client $m$ and then aggregated as a single global model:
\begin{equation}
\begin{split}
    f^g_i &= F(x_i; \theta^{g(t)}_{F, m}), \hat{y_i}= C(f^g_i; \theta_C), \\
    \mathcal{L}^{g(t)}_m &= \mathbb{E}_{(x_i, y_i) \sim \mathcal{D}_m} J (\hat{y}_i, y_i), \\
\end{split}
\end{equation}
where $f^g$ is the extracted global feature. $\mathcal{L}^g(t)$ is calculated by the loss function $J$ at time step $t$ and updates to receive $\theta^{g(t+1)}_{F, m}$. In conventional FL, FedAvg \cite{smith2017federated} considers the global model by averaging all trained global models from clients. However, under the real-world FL setting, client data is collected from different environments, scenarios, or applications. The data distribution $p(X, Y)_m$ from each client presents diversely, which results in poor generalization if treating them equally. To deal with the client imbalanced issue, we propose adaptive aggregation for a fairer aggregated proportion. 

\begin{algorithm}[!t]
    \begin{algorithmic}[1]
        \small
        
        \STATE \textbf{Server}:
        \STATE Initialize $\theta^g_F$, $\theta^l_{F, {\{1:M\}}}$, $\theta_C$, $\theta_{D, {\{1:M\}}}$, $\lambda$
        \STATE
        \FOR{$t = 1~\text{to}~T$}
        \STATE $S_t$ $\leftarrow$ randomly select $m$ clients
        \FOR{$m$ $\in$ $S_t$}
        \STATE $\theta_{F_m}^{g(t+1)}$, $\theta_{C_m}^{(t+1)}$, $\mathcal{L}^{g(t+1)}_m$ $\leftarrow$ Client($m$, $\theta_F^{g(t)}$, $\theta_C^{(t)}$)
        \ENDFOR
        \STATE
        \STATE $\triangleright$ Adaptive aggregation by adaptive $q$
        \STATE $q^{(t+1)} = q^t + \eta_q \frac{\sigma(\mathcal{L}^{g(t+1)})-\sigma(\mathcal{L}^{g(t)})}{(\sigma(\mathcal{L}^{g(t+1)})+\sigma(\mathcal{L}^{g(t)}))/2}$
        \STATE $\lambda_m = \frac{(\mathcal{L}^{g(t+1)}_m)^{q(t+1)}}{\sum_{i=1}^{M} (\mathcal{L}^{g(t+1)}_i)^{q(t+1)}}$
        \STATE $\{\theta_{F}^{g(t+1)}, \theta_{C}^{g(t+1)}\} = \sum_{i=1}^{M} \lambda_i \{\theta_{F, i}^{g(t+1)}, \theta_{C, i}^{g(t+1)}\}$
        \ENDFOR
        \STATE
        \STATE \textbf{Client}($m$, $\theta^g_F$, $\theta_C$):
        \STATE $\theta^g_{F, m}, \theta_{C, m}$ $\leftarrow$ $\theta^g_F, \theta_C$
        \FOR{$r = 1~\text{to}~R$}
        \STATE $\mathcal{B}$ $\leftarrow$ batches ($\mathcal{D}_m$)
        \FOR{batch ($x$, $y$) $\in$ $\mathcal{B}$}
        \STATE $\triangleright$ Run with the global model
        \STATE $\hat{y}^g = C(F(x;\theta^g_{F, m});\theta_{C, m})$
        \STATE $\mathcal{L}^g_m = J(\hat{y}^g, y)$
        \STATE $\theta^g_{F, m} = \theta^g_{F, m} - \eta \nabla_{\theta^g_{F, m}} \mathcal{L}^g_m$
        \STATE $\theta^g_{C, m} = \theta^g_{C, m} - \eta \nabla_{\theta^g_{C, m}} \mathcal{L}^g_m$
        \STATE
        \STATE $\triangleright$ Run with the local model
        \STATE $\hat{y}^l = C(F(x;\theta^l_{F, m});\theta_C)$
        \STATE $\mathcal{L}^l_m = J(\hat{y}^l, y)$
        \STATE $\mathcal{L}_{R, m} = \log(1-D(f^l; \theta_{D, m}))$ $\triangleright$ Update also with $D$
        \STATE $\theta^l_{F_m}$ $\leftarrow$ $\theta^l_{F_m} - \eta \nabla_{\theta^l_{F_m}} (\beta \mathcal{L}^l_m + (1-\beta) \mathcal{L}_{R, m})$
        \STATE
        \STATE $\triangleright$ Update $\theta_{D, m}$ with $f^g$ as true and $f^l$ as false
        \STATE $f^g, f^l = F(x;\theta^g_F), F(x;\theta^l_{F, m})$ 
        \STATE $\mathcal{L}_{D, m} = \log(1-D(f^g; \theta_{D, m})) + \log(D(f^l; \theta_{D, m}))$
        \STATE $\theta_{D, m}$ $\leftarrow$ $\theta_{D, m} - \eta \nabla_{\theta_{D, m}} \mathcal{L}_{D, m}$
        \ENDFOR
        \ENDFOR
        \STATE return $\theta^g_{F, m}$, $\theta_{C, m}$, $\mathcal{L}^g_m$ to server
    \end{algorithmic}
    \caption{GRP-FED, $\eta$: learning rate, $J$: loss function}
    \label{algo:GRP-FED}
\end{algorithm}

\vspace{0.5ex} \textbf{Adaptive Aggregation~}
q-FFL \cite{li2019fair} adopts a constant power $q$ that tunes the amount of fairness. However, the training process for FL can be dynamic over distinct clients, where a fixed power of loss is difficult to satisfy the expected fairness under all situations. To overcome this issue, we present adaptive aggregation and consider a dynamic $q$ to adaptively adjusts for better fairness. We treat fairness as the standard deviation ($\sigma$) of the global training loss $\mathcal{L}^g$ in all clients. If $\sigma$ is high, the global training loss is quite different and the global model may suffer from client imbalance. Therefore, we adjust the loss of power $q$:
\begin{equation}
\label{eq:q}
    q^{(t+1)} = q^t + \eta_q \frac{\sigma(\mathcal{L}^{g(t+1)})-\sigma(\mathcal{L}^{g(t)})}{(\sigma(\mathcal{L}^{g(t+1)})+\sigma(\mathcal{L}^{g(t)}))/2}.
\end{equation}
Otherwise, if the fairness becomes relatively fairer, we should decrease $q$ for more robust training. Finally, we acquire the new global model by aggregating all trained global models, weighted by the global training loss $\mathcal{L}^g$ and the adaptive power $q$:
\begin{equation}
\begin{split}
    \lambda_m &= \frac{(\mathcal{L}^{g(t+1)}_m)^{q(t+1)}}{\sum_{i=1}^{M} (\mathcal{L}^{g(t+1)}_i)^{q(t+1)}}, \\
    \{\theta_{F}^{g(t+1)}, \theta_{C}^{g(t+1)}\} &= \sum_{i=1}^{M} \lambda_i \{\theta_{F, i}^{g(t+1)}, \theta_{C, i}^{g(t+1)}\}.
\end{split}
\end{equation}
In this way, we can adaptively adjust to satisfy the dynamic fairness during the global training by considering the standard deviation of the global training loss over all clients.

\subsection{Local Personalization}
\vspace{0.5ex} \textbf{Local Training~}
Apart from a single global model, since each client is collected from different sources and under various usages, local models that support personalization are also crucial. The local training is to train each local model only with the data in client $m$ for personalization:
\begin{equation}
\begin{split}
    f^l_i &= F(x_i; \theta^{l(t)}_{F, m}), \hat{y_i}= C(f^l_i; \theta_C), \\
    \mathcal{L}^{l(t)}_m &= \mathbb{E}_{(x_i, y_i) \sim \mathcal{D}_m} J (\hat{y}_i, y_i), \\
\end{split}
\end{equation}
where $f^l$ is the extracted personalized feature. Similar to the global training, $\theta^{l(t+1)}_{F, m}$ is updated by $\mathcal{L}^l$ from $J$. Thereby, we personalize the local feature $f^l$ in the specific client. Note that we fix the classifier $C$ with $\theta_{C, m}$ during the local training for a personalized local feature distribution. 

\vspace{0.5ex} \textbf{Global-Regularized Discriminator ($D$)~}
After the local training, we can have the personalized feature $f^l$. However, since under FL, the client data distribution $p(X, Y)_m$ is far from global $p(X, Y)$, the learned $f^l$ may be just overfitting on that client but suffers from poor generalization for the global scenario. To mitigate this overfitting issue, we introduce Global-Regularized Discriminator ($D$). Each client $m$ maintains its own $D$ parameterized by $\theta_{D, m}$, which serves as a binary classifier to distinguish an extracted feature $f$ is from the global or the local feature extractor. We make the global feature $f^g$ by $\theta^g_F$ as the true case and the local feature $f^l$ by $\theta^l_{F, m}$ as the false case, and train $D_m$ as following:
\begin{equation}
\begin{split}
\notag
    f^g_i &= F(x_i; \theta^g_F), f^l_i = F(x_i; \theta^l_{F, m}) \\
    \mathcal{L}_{D, m} &= \mathbb{E}_{x_i \sim \mathcal{D}_m} \log(1-D(f^g_i; \theta_{D, m})) + \log(D(f^l_i; \theta_{D, m})),
\end{split}
\end{equation}
where $\theta^g_F$ and $\theta^l_{F, m}$ are fixed, and only $\theta_{D_m}$ is updated during the $D_m$ training. With the help of $D_m$, the local feature extractor $\theta^g_{F, m}$ can be regularized to prevent overfitting:
\begin{equation}
    \mathcal{L}_{R, m} = \mathbb{E}_{x_i \sim \mathcal{D}_m} \log(1-D(f^l_i; \theta_{D, m})).
\end{equation}
This time, $\theta_{D_m}$ should be freeze and $\theta^l_{F, m}$ is optimized to fool the discriminator $D_m$. By updating the local training along with the global-regularized discriminator, the local feature extractor learns to personalize and imitate the global feature distribution, which can avoid client overfitting.

\begin{table*}[t]
\scriptsize \centering
    \begin{tabular}{cccccccccc}
        \toprule
        ~ & \multicolumn{4}{c}{\textbf{MIT-BIH}} & ~ & \multicolumn{4}{c}{\textbf{CIFAR-10}} \\
        \cmidrule{2-5} \cmidrule{7-10}
        Method & Global Test & Local Test & Personalization & Generalization & ~ & Global Test & Local Test & Personalization & Generalization \\
        \midrule
        Local & 0.075$\pm$0.009 & 0.140$\pm$0.001 & 0.933$\pm$0.005 & 0.076$\pm$0.001& & 0.214$\pm$0.022 &0.290$\pm$0.005 &0.396$\pm$0.009 &0.235$\pm$0.003  \\
        FedAvg & 0.334$\pm$0.046 & 0.407$\pm$0.044 & 0.684$\pm$0.011 & 0.334$\pm$0.046& & 0.462$\pm$0.006 &0.482$\pm$0.004 &0.518$\pm$0.001 &0.462$\pm$0.006\\
        AFL & 0.506$\pm$0.018 &0.503$\pm$0.023 &0.606$\pm$0.042 &0.506$\pm$0.018 & & 0.495$\pm$0.004 &0.496$\pm$0.006 & 0.510$\pm$0.009 & 0.495$\pm$0.004 \\
        q-FFL & 0.551$\pm$0.034 &0.534$\pm$0.006 &0.602$\pm$0.033 &\textbf{0.551$\pm$0.001} & & 0.563$\pm$0.003 &0.530$\pm$0.007 &0.510$\pm$0.009 & \textbf{0.563$\pm$0.003}  \\
        per-FedAvg & 0.378$\pm$0.030 & 0.424$\pm$0.022 & 0.799$\pm$0.004 & 0.310$\pm$0.024& &0.525$\pm$0.021 &0.490$\pm$0.014 &0.550$\pm$0.012 &0.453$\pm$0.017 \\
        pFedMe & 0.290$\pm$0.011 & 0.288$\pm$0.012 & 0.850$\pm$0.006 & 0.178$\pm$0.001& & 0.406$\pm$0.010 &0.414$\pm$0.007 &0.503$\pm$0.014 &0.356$\pm$0.010\\
        LG-FedAvg & 0.343$\pm$0.034 & 0.286$\pm$0.010 & \textbf{0.964$\pm$0.007} & 0.169$\pm$0.006& &0.503$\pm$0.025 &0.499$\pm$0.015 & \textbf{0.676$\pm$0.017} &0.403$\pm$0.014 \\
        GRP-FED & \textbf{0.569$\pm$0.004} & \textbf{0.553$\pm$0.011} & 0.864$\pm$0.022  &0.424$\pm$0.007 &   &\textbf{0.578$\pm$0.001} &\textbf{0.552$\pm$0.010}  &0.611$\pm$0.010  &0.516$\pm$0.007  \\
    \bottomrule
    \end{tabular}
    \vspace{-1ex}
    \caption{The quantitative results of our GRP-FED and baselines in the global test ($T_g$) and the local test ($T_l$), including the personalization test ($T_p$) and the generalization test ($T_r$), on both real-world MIT-BIH and synthesis CIFAR-10 datasets.}
    \vspace{-3ex}
    \label{table:Result}
\end{table*}

\subsection{Learning of GRP-FED and Inference}
The learning process of GRP-FED is presented in Algo.~\ref{algo:GRP-FED}. For each round $t$, to make the federated learning stable, we follow \cite{smith2017federated} that randomly selects $m$ clients as $S_t$ for training. At first, the server copies global feature extractor $\theta^g_F$ and classifier $\theta_C$ to the clients for independent federated training. Both $\theta^g_F$ and $\theta^C$ are trained through data from all clients during the global training. For the local training, the local feature extractor $\theta^l_F$ is only trained by the client data and updated from the $\mathcal{L}^l_m$ to do personalization on client $m$. Also, $\theta^l_F$ is jointly trained from $\mathcal{L}_{R, m}$ to prevent overfitting. The global-regularized discriminator ($D$) $\theta_{D, m}$ then updates from $\mathcal{L}_{D, m}$ by discriminating an feature is extracted from the local or global model, where $\beta$ is the weight of loss between $\mathcal{L}^l_m$ and $\mathcal{L}_{R, m}$.

After returning all trained $\theta^g_{F, m}$, $\theta_{C, m}$, and global training loss $\mathcal{L}^g_m$ from each client $m$, we aggregate $\theta^g_{F, m}$ and $\theta^g_{C, m}$ as the new global model over the aggregated weight $\lambda$. $\lambda$ is updated from the adaptive power $q$ and the global training loss $\mathcal{L}^g$ to force investigating different proportions of clients. In total, the entire training loss $\mathcal{L}_\text{T}$ of GRP-FED is:
\begin{equation}
\notag \scriptsize
    \mathcal{L}_\text{T} =\sum_{m \in S_t} \mathcal{L}^g_m +\underbrace{\sum_{m \in S_t}(\beta \mathcal{L}^l_m + (1-\beta) \mathcal{L}_{R, m})}_{\text{Local Personalization}}+\underbrace{\sum_{m \in S_t} \mathcal{L}_{D, m}}_{\text{Discriminator}}.
\end{equation}

\vspace{0.5ex} \textbf{Inference~}
During inference, given an example $x'$, we consider two testing types for both local and global scenario:
\begin{itemize}[topsep=0pt, noitemsep, leftmargin=*]
    \item local test: if $x'$ belongs to client $m$, we apply the local model ($\theta^l_{F, m}$) for the best personalization;
    \item global test: otherwise, $x'$ is fed to the global model ($\theta^g_{F}$) as an unknown example from the global distribution. 
\end{itemize}
We also conduct these two types of testing in our experiments to evaluate both global fairness (global test) and local personalization(local test) of our proposed GRP-FED. 

\section{Experiments}
\subsection{Experimental Setting}
\vspace{0.5ex} \textbf{Dataset~}
We evaluate our GRP-FED on two federated classification datasets, real-world MIT-BIH \cite{goldberger2000physiobank}, and synthesis CIFAR-10 \cite{cifar10}. MIT-BIH is an Electrocardiography (ECG) dataset for medical diagnosis, where each fragment belongs to one of 12 arrhythmia classes. There are 46 patients in MIT-BIH, containing different numbers of ECG fragments and presenting various class distributions. We treat each patient as a single client that supports both personalized evaluation for a specific patient and global evaluation over all clients.

We distribute the entire CIFAR-10 dataset to 50 clients as the FL setting. To imitate different client distributions, each client contains different numbers of total data (with $\rho$ decreasing over clients). The class distribution is randomly sampled (with $\tau$ decreasing over the number of each class).  $(\rho, \tau)$ is $(0.7, 0.5)$ that follows the distribution of MIT-BIH.

\vspace{0.5ex} \textbf{Evaluation Metrics~}
Since the class distribution over real-world data is non-uniform, the classic accuracy (\%) cannot reflect the proper performance of the prediction and may ignore those minor classes with less examples. We adopt \textbf{macro-F1}, the mean F1-score of each class, to treat them as the same importance. This evaluation is more suitable under data imbalance. For instance, we care more about those examples with different arrhythmia issues in MIT-BIH. 

\vspace{0.5ex} \textbf{Testing Scenario}
We conduct global usage and local personalization under two testing scenarios:
\begin{itemize}[topsep=0pt, noitemsep, leftmargin=*]
    \item Global Test ($T_g$): the global model predicts the entire testing set to evaluate the fairness over global distribution;
    \item Local test ($T_l$): we consider both aspects of local personalization ($T_p$) and local generalization ($T_r$) to avoid overfitting. $T_p$ is the mean performance of local models under their clients; $T_r$ calculates from the mean macro-F1 score of each local model in the global test. Concerning both personalization and overfitting, the overall performance of the local test ($T_l$) is computed as:
    \begin{equation}
        T_l = \frac{2*T_p*T_r}{T_p+T_r}.
    \end{equation}
\end{itemize}

\vspace{0.5ex} \textbf{Baselines~}
We compare against various FL methods:
\begin{itemize}[topsep=0pt, noitemsep, leftmargin=*]
    \item Global-only: FedAvg \cite{smith2017federated}, q-FFL \cite{li2019fair}, and AFL \cite{DBLP:conf/icml/MohriSS19};
    \item Local-only: Local and LG-FedAvg \cite{liang2020think};
    \item Global-Local: pFedMe \cite{DBLP:conf/nips/DinhTN20} and per-FedAvg \cite{fallah2020personalized}.
\end{itemize}
For global-only methods, the global model evaluates under each client to perform the local test. Following LG-FedAvg \cite{liang2020think}, we ensemble results from all local models as the global output for local-only algorithms. With GRP-FED or global-local frameworks, we apply the global model for the global test and local models for the local test.

\vspace{0.5ex} \textbf{Implementation Detail~}
As the classification task, we apply the cross-entropy loss for the loss function $J$. We adopt 5-layer 1D ResNet \cite{he2016res} to process ECG under MIT-BIH and ResNet-30 under CIFAR10 as the feature extractor $F$. The classifier $C$ is a 2-layer fully-connected (FC) that projects the feature into class prediction. The global-regularized discriminator $D$ is also 2-layer FC but projects to binary indication for the true/false discrimination. We set the local epoch $R=5$ and the batch size 64. SGD optimizes all parameters with a learning rate ($\eta$) 5e-3, $q$ adjusting rate ($\eta_q$ in Eq.~\ref{eq:q}) 0.5, momentum 0.9. The initial loss power $q$ is 10, which is the same as q-FFL. 

\subsection{Quantitative Results}
\vspace{0.5ex} \textbf{Global Test~}
Table~\ref{table:Result} shows the results of GRP-FED and baselines on both real-world MIT-BIH and synthesis CIFAR-10 datasets. The global test ($T_g$) is to evaluate the fairness of the global model over the entire testing set. It shows that our GRP-FED achieves the highest macro-F1 score on both MIT-BIH ($56.9\%$) and CIFAR-10 ($57.8\%$). Since the proposed adaptive aggregation adjusts the power of loss according to the dynamic fairness, it gains a significant improvement under $T_g$ and achieves better global fairness.

\vspace{0.5ex} \textbf{Local Test~}
Local personalization is essential when regarding a specific client under the FL setting. At first, LG-FedAvg \cite{liang2020think} performs the best in the local personalization test (96.4\% $T_p$) on MIT-BIH. However, since the local features are merely learned from the client, they are easily overfitting and result in poor generalization in the local generalization test (16.9\% $T_r$). q-FFL has the highest 55.1\% $T_r$ but presents lower 60.2\% $T_p$ without personalization. With the global-regularized discriminator, local models in our GRP-FED can extract personalized features but avoid overfitting. We surpass all baselines in the overall local test (55.3\% $T_l$) with a comparable 86.4\% $T_p$ and 42.4\% $T_r$. A similar trend can be found on CIFAR-10. Our GRP-FED achieves the highest 55.2\% $T_l$ and strikes the most appropriate balance between personalization (61.1\% $T_p$) and generalization (51.6\% $T_r$).

\subsection{Ablation Study}
\begin{figure}[t]
    \centering
    \includegraphics[width=\linewidth]{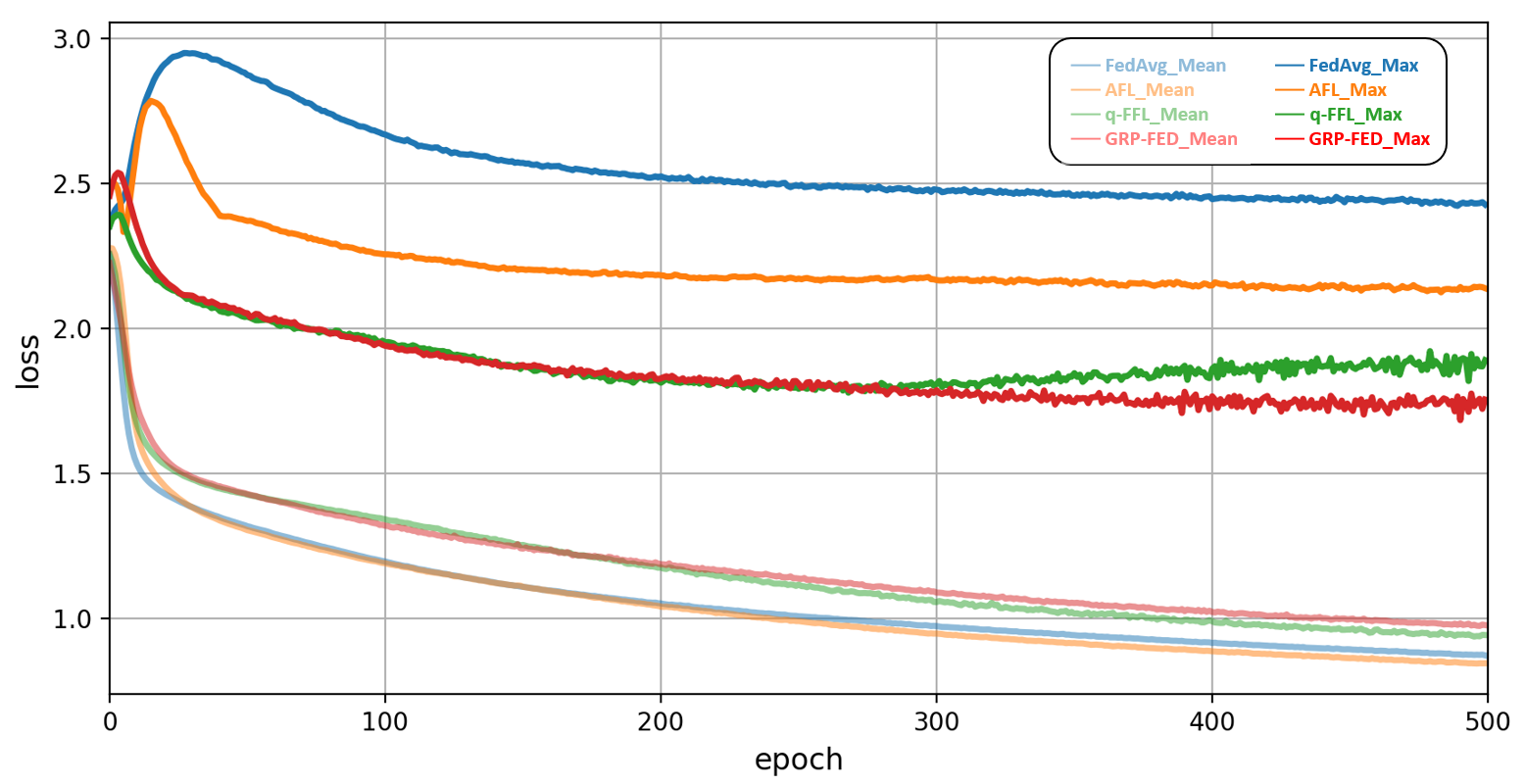}
    \vspace{-4ex}
    \caption{Learning curve of the \textit{mean}/\textit{max} global training loss.}
    \label{fig:loss_curve}
    \vspace{-3ex}
\end{figure}

\vspace{0.5ex} \textbf{Is the Global Model Actually Fair?~}
To ensure the global model is actually fair, we plot the learning curve of the \textit{mean} and \textit{max} global training loss in Fig.~\ref{fig:loss_curve}. Basically, all methods have a relatively low \textit{mean} training loss during the global training. We can investigate the global fairness through the \textit{max} global training loss. FedAvg treats each client equally and sacrifices those minor classes, resulting in a high \textit{max} global training loss in the end. The adversarial aggregation in AFL is not fair enough and still remains high \textit{max} training loss. q-FFL adopts a constnat loss power $q$ to tune the amount of fairness. While, a fixed power of loss cannot satisfy all fairness situations and instead increases the \textit{max} training loss at last. Our adaptive aggregation considers a dynamic $q$ that can adaptively adjust, which keeps decreasing the \textit{max} training loss as the fairer global model.

\begin{figure}[t]
    \centering
    \includegraphics[width=\linewidth]{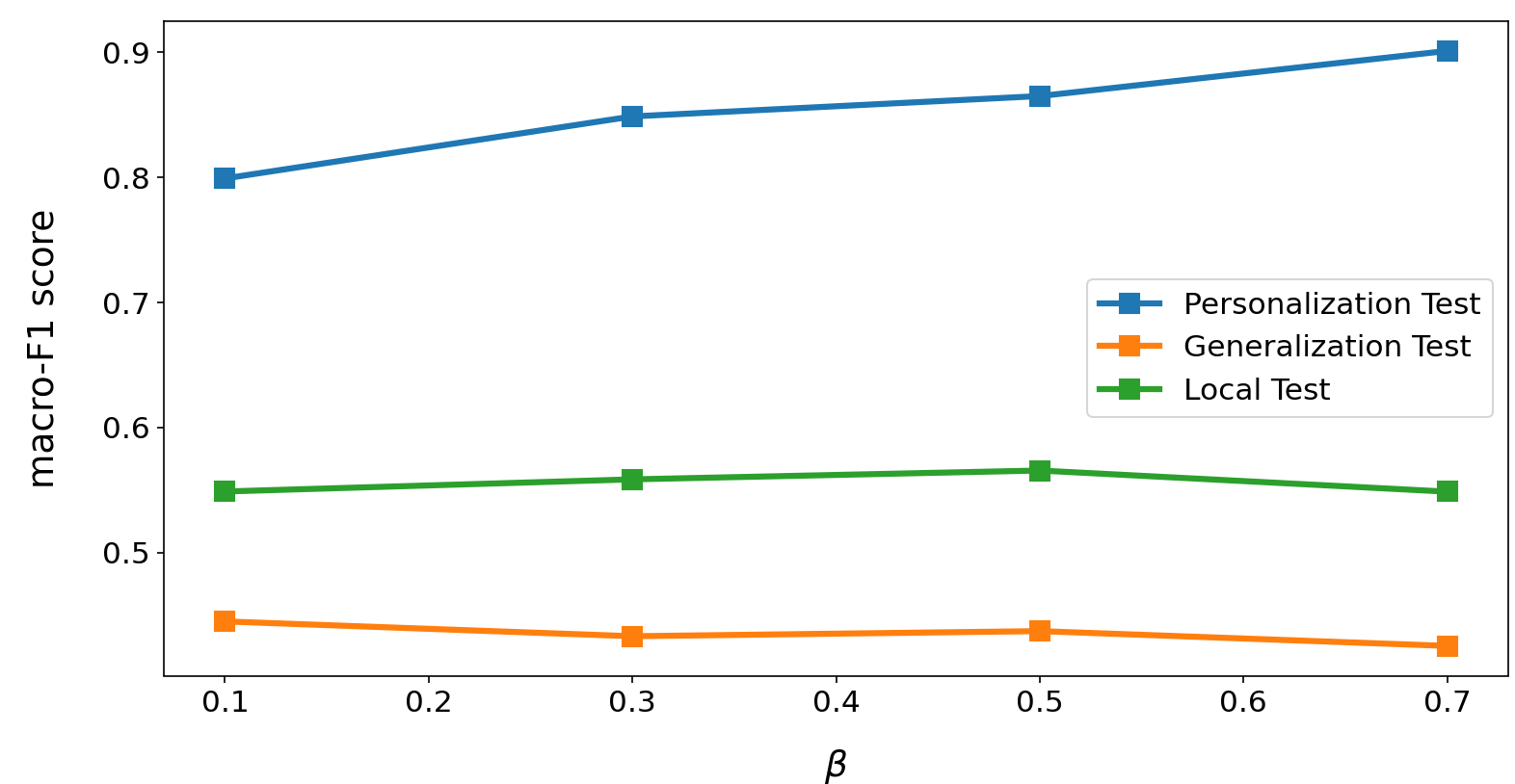}
    \vspace{-4ex}
    \caption{The trade-off between the personalization ($T_p$) and generalization ($T_r$) in local test ($T_l$) under different loss weight $\beta$.}
    \label{fig:parameter_sensitivity}
    \vspace{-1ex}
\end{figure}

\vspace{0.5ex} \textbf{How $\beta$ affects the local models?~}
We adopt $\beta$ to control the weight of loss between the personalization by the local training and the generalization by the global-regularized discriminator. Fig.~\ref{fig:parameter_sensitivity} illustrates the effect of $\beta$ on MIT-BIH during the local personalization. There is a trade-off between local personalization ($T_p$) and local generalization ($T_r$). When $\beta$ gets larger, we treat the personalization as more important and improve $T_p$ but hurt $T_r$. On the other hand, $T_r$ increases when $T_p$ decreases if we consider the local feature should be more generalized with lower $\beta$. $\beta=0.5$ leads to the best local test ($T_l$) in our GRP-FED.

\begin{figure}[t]
    \begin{minipage}[htbp]{.44\linewidth}
        \centering
        \includegraphics[width=\linewidth]{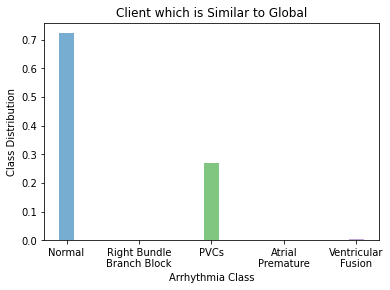}
    \end{minipage}
    \begin{minipage}[htbp]{\linewidth}
        \resizebox{0.55\linewidth}{10mm}{
        \begin{tabular}{c|ccc}
            \toprule
            Method      & Global Test & Local Test & Personalization  \\
            \midrule
            FedAvg      & 0.996 & 0.498 & 0.996 \\
            q-FFL       & 0.718 & 0.464 & 0.578\\
            per-FedAvg  & 0.990 & 0.519 & 0.999 \\
            pFedMe      & 0.992 & 0.379 & 0.998 \\
            LG-FedAvg   & 0.999 & 0.380 & \textbf{1.000} \\
            GRP-FED     & \textbf{0.997} & \textbf{0.657} & 0.998 \\
            \bottomrule
        \end{tabular}
        }
    \end{minipage}
    \begin{minipage}[htbp]{0.44\linewidth}
        \centering
        \includegraphics[width=\linewidth]{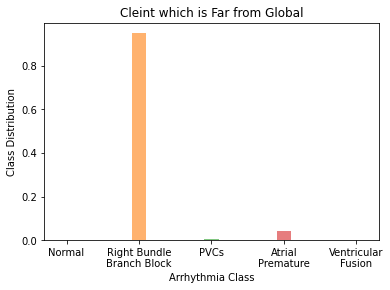}
    \end{minipage}
    \begin{minipage}[htbp]{\linewidth}
        \resizebox{0.55\linewidth}{10mm}{
        \begin{tabular}{c|ccc}
            \toprule
            Method      & Global Test & Local Test & Personalization  \\
            \midrule
            FedAvg      & 0.219 & 0.262 & 0.219 \\
            q-FFL       & 0.328 & 0.325 & 0.328  \\
            per-FedAvg  & 0.291 & 0.351 & 0.435 \\
            pFedMe      & 0.152 & 0.274 & 0.492  \\
            LG-FedAvg   & 0.269 & 0.178 & \textbf{0.975}  \\
            GRP-FED     & \textbf{0.454} & \textbf{0.491} & 0.493\\
            \bottomrule
        \end{tabular}
        }
    \end{minipage}
    
    \vspace{-1ex}
    \caption{The performance of global test and local test in two clients (upper: similar to, lower: far from the global distribution).}
    \vspace{-3ex}
    \label{fig:case_study}
\end{figure}

\vspace{0.5ex} \textbf{Case Study~}
Fig.~\ref{fig:case_study} visualizes the performance of two clients on MIT-BIH. Since the upper client is similar to the global class distribution, all methods perform well under both the global test ($T_g$) and local personalization test ($T_p$). The trained local model on this client also performs well in the overall local test ($T_l$). However, in the lower client, which presents a distinct class distribution, LG-FedAvg is entirely overfitting and results in high $T_p$ but terrible $T_l$. By contrast, our GRP-FED still outperforms the remaining baselines in $T_p$ and achieves the highest $T_l$ with the help of the global-regularized discriminator. Moreover, GRP-FED considers the dynamic fairness from different client distributions and leads to the best $T_g$ as a fairer global model.

\section{Conclusion}
In this paper, we introduce Global-Regularized Personalization (GRP-FED) to address the client imbalance issue under federated learning (FL). GRP-FED consists of a global model and local models for each client. The global model considers the dynamic fairness and investigates different proportions of clients with the adaptive aggregation. The local models do personalization by the local training, and the proposed global-regularized discriminator can prevent the overfitting issue. Extensive results show that our GRP-FED outperforms baselines under both global and local scenarios on real-world MIT-BIH and synthesis CIFAR-10 datasets.

\bibliography{main}
\bibliographystyle{icml2021}

\end{document}